\documentclass{article}

\usepackage{microtype}
\usepackage{graphicx}
\usepackage{subcaption}
\usepackage{booktabs} 

\usepackage{hyperref}

\usepackage{amsfonts}
\usepackage{amsmath}
\usepackage[utf8]{inputenc}
\usepackage{tikz}
\usepackage{pgfplots}
\usepackage[group-separator={,}]{siunitx}
\usepackage{xcolor}
\usepackage{xspace}

\pgfplotsset{compat=newest}

\usetikzlibrary{backgrounds}
\usetikzlibrary{calc}
\usetikzlibrary{external}
\usetikzlibrary{fit}
\usetikzlibrary{positioning}
\usetikzlibrary{shapes}

\tikzexternaldisable

\usepackage[accepted]{icml2018}

\icmltitlerunning{Synthesizing Programs for Images using Reinforced Adversarial Learning}

\newcommand{\xx}{{\mathbf x}}
\newcommand{\yy}{{\mathbf y}}
\newcommand{\xxsub}[1]{\mathbf{x}_{\mbox{\scriptsize #1}}}
\newcommand{\norm}[1]{\left\lVert #1 \right\rVert}

\DeclareMathOperator{\Lip}{Lip}

\newcommand*{\eg}{{\em e.g.}\@\xspace}
\newcommand*{\ie}{{\em i.e.}\@\xspace}

\makeatletter
\newcommand*{\etc}{%
    \@ifnextchar{.}%
        {etc}%
        {etc.\@\xspace}%
}
\makeatother

\makeatletter
\newcommand{\todo}[1][]{\@latex@warning{TODO #1}}
\makeatother

\newcommand{\fig}[1]{Figure~\ref{fig:#1}}
\newcommand{\sect}[1]{Section~\ref{sect:#1}}

\newcommand{\eq}[1]{(\ref{eq:#1})}

\newcommand{\dataset}[1]{\textsc{#1}}

\newlength\figureheight
\newlength\figurewidth

\newif\ifshowplots
\showplotstrue 

\definecolor{dmyellow}{HTML}{F8E46C}
\definecolor{dmlightblue}{HTML}{7DC3D2}
\definecolor{dmred}{HTML}{EF5A58}
\definecolor{dmpurple}{HTML}{DA81B3}
\definecolor{dmgreen}{HTML}{BCD26F}

\usepackage{tikz}
\usetikzlibrary{shapes, arrows, fit}
\usetikzlibrary{positioning,shadows}
\usetikzlibrary{decorations.pathreplacing}
\definecolor{inodefill}{rgb} {0.00,0.53,0.88}
\definecolor{inodedraw}{rgb} {1.00,1.00,1.00}
\definecolor{pnodefill}{rgb} {1.00,1.00,1.00}
\definecolor{onodedraw}{rgb} {0.00,0.00,0.00}
\definecolor{onodefill}{rgb} {1.00,1.00,1.00}
\definecolor{pnodedraw}{rgb} {0.00,0.00,0.00}
\definecolor{mnodefill}{rgb} {0.00,0.53,0.88}

\input{figures/teaser/spiral}

\begin{document}

\twocolumn[
\icmltitle{Synthesizing Programs for Images using Reinforced Adversarial Learning}



\icmlsetsymbol{equal}{*}

\begin{icmlauthorlist}
\icmlauthor{Yaroslav Ganin}{mila}
\icmlauthor{Tejas Kulkarni}{dm}
\icmlauthor{Igor Babuschkin}{dm}
\icmlauthor{S. M. Ali Eslami}{dm}
\icmlauthor{Oriol Vinyals}{dm}
\end{icmlauthorlist}

\icmlaffiliation{mila}{Montreal Institute for Learning Algorithms, Montr{\'e}al, Canada}
\icmlaffiliation{dm}{DeepMind, London, United Kingdom}

\icmlcorrespondingauthor{Yaroslav Ganin}{yaroslav.ganin@gmail.com}

\icmlkeywords{Adversarial, Reinforcement Learning, Visual, Program Synthesis}

\vskip 0.3in
]



\printAffiliationsAndNotice{Work done while YG was an intern at DeepMind.}  

\begin{abstract}
    Advances in deep generative networks have led to impressive results in recent years. Nevertheless, such models can often waste their capacity on the minutiae of datasets, presumably due to weak inductive biases in their decoders. This is where graphics engines may come in handy since they abstract away low-level details and represent images as high-level programs. Current methods that combine deep learning and renderers are limited by hand-crafted likelihood or distance functions, a need for large amounts of supervision, or difficulties in scaling their inference algorithms to richer datasets. To mitigate these issues, we present SPIRAL, an adversarially trained agent that generates a program which is executed by a graphics engine to interpret and sample images. The goal of this agent is to fool a discriminator network that distinguishes between real and rendered data, trained with a distributed reinforcement learning setup without any supervision. A surprising finding is that using the discriminator's output as a reward signal is the key to allow the agent to make meaningful progress at matching the desired output rendering. To the best of our knowledge, this is the first demonstration of an end-to-end, unsupervised and adversarial inverse graphics agent on challenging real world (\dataset{MNIST}, \dataset{Omniglot}, \dataset{CelebA}) and synthetic 3D datasets. A video of the agent can be found at \url{https://youtu.be/iSyvwAwa7vk}.
\end{abstract}

\section{Introduction}

\begin{figure}[t!]
    \centering
    \input{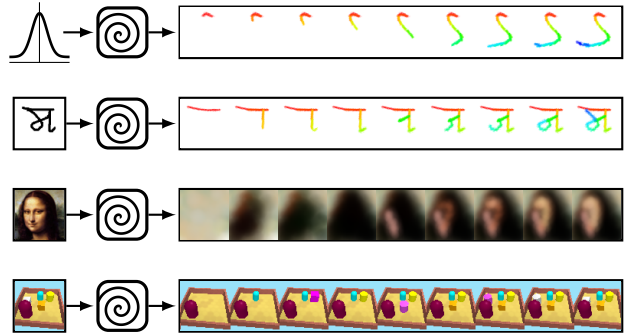}
    \caption{\textbf{SPIRAL} takes as input either random noise or images and iteratively produces plausible samples or reconstructions via graphics program synthesis. The first row depicts an unconditional run given random noise. The second, third and fourth rows depict conditional execution given an image with a handwritten character, the Mona Lisa, and objects arranged in a 3D scene.\vspace{-1mm}}
    \label{fig:teaser}
\end{figure}

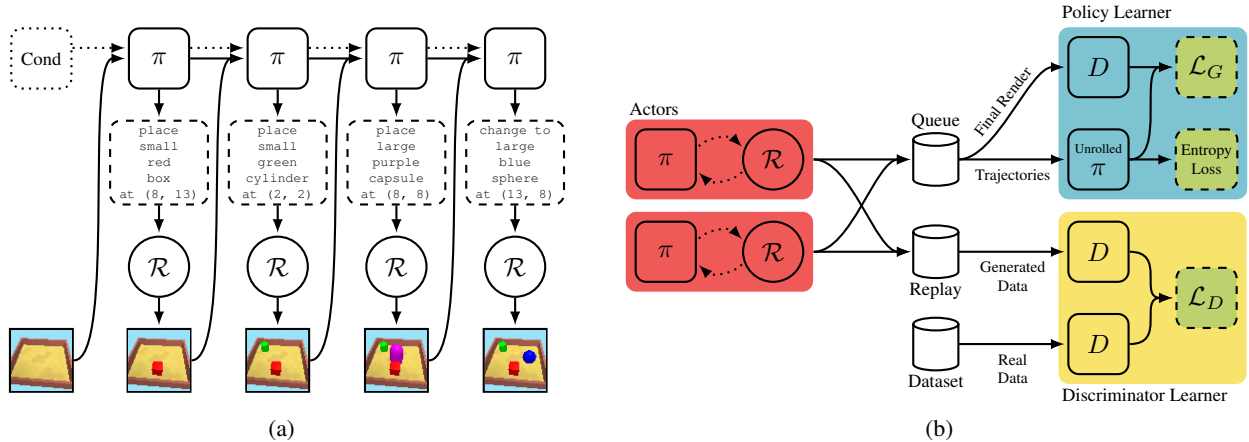
\begin{figure*}[t!]
    \centering
    \begin{subfigure}[b]{0.49\textwidth}
        \centering
        \colorlet{linecolor}{black}
\begin{tikzpicture}[
  every text node part/.style={align=center},
  linecolor, text=black,
  node distance=4mm,
  pinode/.style={
    align=center,
    rectangle,minimum size=0.8cm,rounded corners,
    inner sep=2pt,
    thick,draw=linecolor,
    font=\normalsize},
  anode/.style={
    align=center,
    rectangle,minimum size=0.8cm,minimum width=1.3cm,rounded corners,
    inner sep=2pt,
    thick,draw=linecolor,dashed,
    fill=none,
    font=\ttfamily\fontsize{5}{6}\selectfont},
  rnode/.style={
    align=center,
    circle,minimum size=0.8cm,
    inner sep=2pt,
    thick,draw=linecolor,
    font=\normalsize},
  cnode/.style={
    align=center,
    inner sep=0.5pt,
    thick,draw=linecolor}]
\matrix[row sep=4mm,column sep=2.5mm] {
    \node (cond) [pinode,dotted,font=\scriptsize] {Cond}; &
    \node (pi_1) [pinode] {$ \pi $}; & 
    \node (pi_2) [pinode] {$ \pi $}; &
    \node (pi_3) [pinode] {$ \pi $}; &
    \node (pi_4) [pinode] {$ \pi $}; \\

    \node [minimum width=1.3cm] {}; &
    \node (A_1) [anode] {place\\small\\red\\box\\ at (8, 13)}; &
    \node (A_2) [anode] {place\\small\\green\\cylinder\\at (2, 2)}; &
    \node (A_3) [anode] {place\\large\\purple\\capsule\\at (8, 8)}; &
    \node (A_4) [anode] {change to\\large\\blue\\sphere\\at (13, 8)}; \\

    &
    \node (R_1) [rnode] {$ \mathcal{R} $}; &
    \node (R_2) [rnode] {$ \mathcal{R} $}; &
    \node (R_3) [rnode] {$ \mathcal{R} $}; &
    \node (R_4) [rnode] {$ \mathcal{R} $}; \\

    \node (C_0) [cnode] {\pgfimage[height=0.8cm]{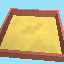}}; &
    \node (C_1) [cnode] {\pgfimage[height=0.8cm]{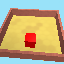}}; &
    \node (C_2) [cnode] {\pgfimage[height=0.8cm]{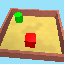}}; &
    \node (C_3) [cnode] {\pgfimage[height=0.8cm]{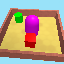}}; &
    \node (C_4) [cnode] {\pgfimage[height=0.8cm]{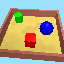}}; \\
};

\path (pi_1) edge[-latex,shorten >=1pt,thick] (A_1);
\path (pi_2) edge[-latex,shorten >=1pt,thick] (A_2);
\path (pi_3) edge[-latex,shorten >=1pt,thick] (A_3);
\path (pi_4) edge[-latex,shorten >=1pt,thick] (A_4);

\path (pi_1) edge[-latex,shorten >=1pt,thick] (pi_2);
\path (pi_2) edge[-latex,shorten >=1pt,thick] (pi_3);
\path (pi_3) edge[-latex,shorten >=1pt,thick] (pi_4);

\path (A_1) edge[-latex,shorten >=1pt,thick] (R_1);
\path (A_2) edge[-latex,shorten >=1pt,thick] (R_2);
\path (A_3) edge[-latex,shorten >=1pt,thick] (R_3);
\path (A_4) edge[-latex,shorten >=1pt,thick] (R_4);

\path (R_1) edge[-latex,shorten >=1pt,thick] (C_1);
\path (R_2) edge[-latex,shorten >=1pt,thick] (C_2);
\path (R_3) edge[-latex,shorten >=1pt,thick] (C_3);
\path (R_4) edge[-latex,shorten >=1pt,thick] (C_4);

\draw[-latex,shorten >=1pt,thick,draw=linecolor] (C_0) .. controls ++(0:1) and ++(0:-1) .. (pi_1);
\draw[-latex,shorten >=1pt,thick,draw=linecolor] (C_1) .. controls ++(0:1) and ++(0:-1) .. (pi_2);
\draw[-latex,shorten >=1pt,thick,draw=linecolor] (C_2) .. controls ++(0:1) and ++(0:-1) .. (pi_3);
\draw[-latex,shorten >=1pt,thick,draw=linecolor] (C_3) .. controls ++(0:1) and ++(0:-1) .. (pi_4);


\draw[-latex,shorten >=1pt,thick,dotted] ($ (cond.east) + (0,4pt) $) -- ($ (pi_1.west) + (0,4pt) $);
\draw[-latex,shorten >=1pt,thick,dotted] ($ (pi_1.east) + (0,4pt) $) -- ($ (pi_2.west) + (0,4pt) $);
\draw[-latex,shorten >=1pt,thick,dotted] ($ (pi_2.east) + (0,4pt) $) -- ($ (pi_3.west) + (0,4pt) $);
\draw[-latex,shorten >=1pt,thick,dotted] ($ (pi_3.east) + (0,4pt) $) -- ($ (pi_4.west) + (0,4pt) $);

\end{tikzpicture}
        \caption{}
        \label{fig:mujoco_actor_diagram}
    \end{subfigure} %
    \hfill
    \begin{subfigure}[b]{0.49\textwidth}
        \centering
        \colorlet{linecolor}{black}
\begin{tikzpicture}[
  every text node part/.style={align=center},
  linecolor, text=black,
  node distance=4mm,
  pinode/.style={
    align=center,
    rectangle,minimum size=0.8cm,rounded corners,
    inner sep=2pt,
    thick,draw=linecolor,
    font=\normalsize},
  dnode/.style={
    align=center,
    cylinder,shape border rotate=90,minimum width=0.6cm,minimum height=0.7cm,shape aspect=1.2,
    anchor=shape center,
    inner sep=2pt,
    thick,draw=linecolor,
    font=\normalsize},
  lnode/.style={
    align=center,
    rectangle,minimum size=0.8cm,rounded corners,
    inner sep=2pt,
    thick,dashed,draw=linecolor,fill=dmgreen,
    font=\normalsize},
  rnode/.style={
    align=center,
    circle,minimum size=0.8cm,
    inner sep=2pt,
    thick,draw=linecolor,
    font=\normalsize}]
\matrix[row sep=4mm,column sep=6mm] {
    & & & & &
    \node (D_1) [pinode] {$ D $}; &
    \node (g_loss) [lnode] {$ \mathcal{L}_G $}; \\

    \node (pi_1) [pinode] {$ \pi $}; & 
    \node (R_1) [rnode] {$ \mathcal{R} $}; &
    \node [minimum width=0.1cm] {}; &
    \node (queue) [dnode] {}; &
    \node [minimum width=0.1cm] {}; &
    \node (unrolled_pi) [pinode] {{\fontsize{5}{6}\selectfont Unrolled}\\[-4pt]$ \pi $}; &
    \node (entropy_loss) [lnode,font=\tiny] {Entropy \\ Loss}; \\

    \node (pi_2) [pinode] {$ \pi $}; & 
    \node (R_2) [rnode] {$ \mathcal{R} $}; & &
    \node (replay) [dnode] {}; & &
    \node (D_2) [pinode] {$ D $}; &
    \node (dummy_1) {}; \\

    & & &
    \node (dataset) [dnode] {}; & &
    \node (D_3) [pinode] {$ D $}; &
    \node (dummy_2) {}; \\
};

\node (d_loss) [lnode] at ($ (dummy_1)!0.5!(dummy_2) $) {$ \mathcal{L}_D $};

\begin{pgfonlayer}{background}
    \node (actor_1) [fit=(pi_1)(R_1),rounded corners,fill=dmred] {};
    \node (actor_2) [fit=(pi_2)(R_2),rounded corners,fill=dmred] {};
    \node (policy_learner) [fit=(D_1)(entropy_loss),rounded corners,fill=dmlightblue] {};
    \node (disc_learner) [fit=(D_2)(D_3)(d_loss),rounded corners,fill=dmyellow] {};
\end{pgfonlayer}

\node [above right=2pt,inner sep=0] at (actor_1.north west) {\scriptsize Actors};
\node [above right=2pt,inner sep=0] at (policy_learner.north west) {\scriptsize Policy Learner};
\node [below right=2pt,inner sep=0] at (disc_learner.south west) {\scriptsize Discriminator Learner};

\node [above=1pt of queue,inner sep=0] {\scriptsize Queue};
\node [below=1pt of replay,inner sep=0] {\scriptsize Replay};
\node [below=1pt of dataset,inner sep=0] {\scriptsize Dataset};

\draw[-latex,shorten <=1pt,shorten >=1pt,thick,dotted] (pi_1.20) to [bend left=45] (R_1.160);
\draw[-latex,shorten <=1pt,shorten >=1pt,thick,dotted] (R_1.200) to [bend left=45] (pi_1.340);

\draw[-latex,shorten <=1pt,shorten >=1pt,thick,dotted] (pi_2.20) to [bend left=45] (R_2.160);
\draw[-latex,shorten <=1pt,shorten >=1pt,thick,dotted] (R_2.200) to [bend left=45] (pi_2.340);

\coordinate (queue_west) at ($ (queue.shape center) - (0.3cm,0) $);
\coordinate (queue_east) at ($ (queue.shape center) + (0.3cm,0) $);
\path (actor_1) edge[-latex,shorten >=1pt,thick] (queue_west);
\draw[-latex,shorten >=1pt,thick,draw=linecolor] (queue_east) .. controls ++(0:0.5) and ++(180:0.5) .. (D_1.west) node [pos=0.52,above,sloped,font=\tiny] {Final Render};
\draw[-latex,shorten >=1pt,thick] (queue_east) -- (unrolled_pi) node [midway,below] {\tiny Trajectories};
\path (D_1) edge[-latex,shorten >=1pt,thick] (g_loss);
\path (unrolled_pi) edge[-latex,shorten >=1pt,thick] (entropy_loss);

\draw[-latex,shorten >=1pt,thick,draw=linecolor] (unrolled_pi.east) .. controls ++(0:0.5) and ++(180:0.5) .. (g_loss.west);

\coordinate (replay_west) at ($ (replay.shape center) - (0.3cm,0) $);
\coordinate (replay_east) at ($ (replay.shape center) + (0.3cm,0) $);
\path (actor_2) edge[-latex,shorten >=1pt,thick] (replay_west);
\draw[-latex,shorten >=1pt,thick] (replay_east) -- (D_2) node[midway,below,font=\tiny] {Generated\\Data};

\coordinate (dataset_east) at ($ (dataset.shape center) + (0.3cm,0) $);
\draw[-latex,shorten >=1pt,thick] (dataset_east) -- (D_3) node[midway,below,font=\tiny] {Real\\Data};

\draw[-latex,shorten >=1pt,thick,draw=linecolor] (D_2.east) .. controls ++(0:0.5) and ++(180:0.5) .. (d_loss.west);
\draw[-latex,shorten >=1pt,thick,draw=linecolor] (D_3.east) .. controls ++(0:0.5) and ++(180:0.5) .. (d_loss.west);

\draw[-latex,shorten >=1pt,thick,draw=linecolor] (actor_1.east) .. controls ++(0:0.5) and ++(180:0.5) .. (replay_west);
\draw[-latex,shorten >=1pt,thick,draw=linecolor] (actor_2.east) .. controls ++(0:0.5) and ++(180:0.5) .. (queue_west);

\end{tikzpicture}
        \caption{}
        \label{fig:distributed_training}
    \end{subfigure}
    \caption{\textbf{The SPIRAL architecture.} \textbf{(a)} An execution trace of the SPIRAL agent. The policy outputs program fragments which are rendered into an image at each step via a graphics engine $ \mathcal{R} $. The agent can make use of these intermediate renders to adjust its policy. The agent only receives a reward in the final step of execution. \textbf{(b)} Distributed training of SPIRAL. A collection of actors (in our experiments, up to 64), asynchronously and continuously produce execution traces. This data, along with a training dataset of ground-truth renderings, are passed to a Wasserstein discriminator on a separate GPU for adversarial training. The discriminator assesses the similarity of the final renderings of the traces to the ground-truth. A separate off-policy GPU learner receives batches of execution traces and trains the agent's parameters via policy-gradients to maximize the reward assigned to them by the discriminator, \ie, to match the distribution of the ground truth dataset.\vspace{-0mm}}
    \label{fig2}
\end{figure*}

Recovering \textit{structured} representations from raw sensations is an ability that humans readily possess and frequently use. Given a picture of a hand-written character, decomposing it into strokes can make it easier to classify or re-imagine that character, and similarly, knowing the underlying layout of a room can aid with planning, navigation and interaction in that room. Furthermore, this structure can be exploited for generalization, rapid learning, and even communication with other agents. It is commonly believed that humans exploit \textit{simulations} to learn this skill \cite{Lake17}. By experimenting with a pen and a piece of paper we learn how our hand movements lead to written characters, and via imagination we learn how architectural layouts manifest themselves in reality.

In the visual domain, inversion of a renderer for the purposes of scene understanding is typically referred to as inverse graphics \cite{Mansinghka13,Kulkarni15b}. Training vision systems using the inverse graphics approach has remained a challenge. Renderers typically expect as input \textit{programs} that have sequential semantics, are composed of discrete symbols (\eg, keystrokes in a CAD program), and are long (tens or hundreds of symbols). Additionally, matching rendered images to real data poses an optimization problem as black-box graphics simulators are not differentiable in general.  

To address these problems, we present a new approach for interpreting and generating images using \emph{Deep Reinforced Adversarial Learning}. In this approach, an adversarially trained agent generates visual programs which are in turn executed by a graphics engine to generate images, either conditioned on data or unconditionally. The agent is rewarded by fooling a discriminator network, and is trained with distributed reinforcement learning without any extra supervision. The discriminator  network itself is trained to distinguish between rendered and real images.

Our contributions are as follows:
\begin{itemize}
    \setlength{\itemsep}{6pt}
    \item An adversarially trained reinforcement learning agent that interprets and generates images in the space of visual programs. Crucially, the architecture of our agent is agnostic both to the semantics of the visual program and to the domain.
    \item Scaling inverse graphics to real world and procedural datasets without the need for labels. In particular, our model discovers pen strokes that give rise to \dataset{MNIST} and \dataset{Omniglot} characters, brush strokes that give rise to celebrity faces, and scene descriptions that, once rendered, reconstruct an image of a 3D scene (\fig{teaser}).
    \item Evidence that utilizing a discriminator's output as the reward signal for reinforcement learning is significantly better at optimizing the pixel error between renderings and data, compared to directly optimizing pixel error.
    \item A showcase of state-of-the-art deep reinforcement learning techniques, which can provide a scaling path for inverse graphics, and could lead to broader implications for program synthesis in future work.
\end{itemize}
\vfill

\section{Related Work}

The idea of inverting simulators to interpret images has been explored extensively in recent years \cite{Nair08,Paysan09,Mansinghka13,Loper14,Kulkarni15b,Jampani15}. Structured object-attribute based `de-rendering' models have been proposed for interpretation of images \cite{Wu17a} and videos \cite{Wu17b}. Concurrent work has explored the use of \emph{Constructive Solid Geometry} primitives for explaining binary images \cite{Sharma17}. \citet{Loper14} proposed the idea of differentiable inverse graphics, which is efficient for optimizing continuous variables but cannot handle discrete variables. Earlier work has also explored using reinforcement learning for automatic generation of single brush strokes \cite{Xie13}. However, scaling these approaches to larger real-world datasets, particularly at test-time, has remained a challenge.

Inferring motor programs for the reconstruction of \dataset{MNIST} digits was first studied in \cite{Nair06}. The generative model is parametrized by two pairs of opposing springs whose stiffness is controlled by a motor program. The training procedure involved starting with a prototype program and its corresponding observation. Random noise was then added to this prototype in order to produce new training examples until the generated distribution stretched to cover the manifold of the training digits. In contrast, our model automatically learns the training curriculum via the discriminator and the same agent is suitable for a range of scene understanding problems, including those in 3D. 

Visual program induction has recently been studied in the context of hand-written characters on the \dataset{Omniglot} dataset \cite{Lake15}. This model achieves impressive performance but requires parses from a hand-crafted algorithm to initialize training and was not demonstrated to generalize beyond hand-written characters. \citet{Ellis17} proposed a visual program induction model to infer \LaTeX programs for diagram understanding. More recently, the \texttt{sketch-rnn} model \cite{Ha17} used sequence-to-sequence learning \cite{Sutskever14} to produce impressive sketches both unconditionally and conditioned on data. However, similar to the aforementioned works and unlike SPIRAL, the model requires supervision in the form of sketches and corresponding sequences of strokes.

In the neural network community, there have been analogous attempts at inferring and learning feed-forward or recurrent procedures for image generation \cite{Lecun15,Hinton06,Goodfellow14,Ackley87,Kingma13,Oord16,Kulkarni15a,Eslami16,Reed17,Gregor15}. These models demonstrate impressive image generation capabilities but generally lack the ability to infer structured representations of images. 

Our approach employs adversarial training techniques, first used for generative modeling \citep{Goodfellow14} and domain adaptation \citep{Ganin15}. Generative Adversarial Networks (GANs) \cite{Goodfellow14} were orignally used for image generation but have now been successfully applied to model audio, text and motor behaviors \cite{Ho16,Merel17}. Perhaps the most interesting extension in our context is their use in domain transfer, where images from one domain (\eg, segmentations) were mapped to another (\eg, pixels). Models such as \texttt{pix2pix} \cite{Isola17}, \texttt{CycleGAN} \cite{Zhu17} and AIGN \cite{Tung17} fall in this category.

The SPIRAL agent builds upon this literature, has minimal hand-crafting in its design, requires no supervision in the form of pairs of programs and corresponding images and, as we demonstrate in the following sections, is applicable across a wide range of domains.

\section{The SPIRAL Agent}

\subsection{Overview}

Our goal is to construct a generative model $ G $ capable of sampling from some target data distribution $ p_d $. To that end, we propose using an external black-box rendering simulator $ \mathcal{R} $ that accepts a sequence of commands $ a = ( a_1, a_2, \ldots, a_N ) $ and transforms them into the domain of interest, \eg, a bitmap. For example, $ \mathcal{R} $ could be a CAD program rendering descriptions of primitives into 3D scenes. Thus, our task is equivalent to recovering a distribution $ p_a $ such that $ p_d \approx \mathcal{R}(p_a) $. We model $ p_a $ with a recurrent neural network $ \pi $ which we call the \emph{policy} network (or, somewhat sloppily, the \emph{agent}). The generation process $ G $, which consists of a policy $ \pi $ and a renderer $ \mathcal{R}$, is illustrated in \fig{mujoco_actor_diagram}.

In order to optimize $ \pi $, we employ the adversarial framework \cite{Goodfellow14}. In this framework, the \emph{generator} (denoted as $ G $) aims to maximally confuse a \emph{discriminator} network $ D $ which is trained to distinguish between the samples drawn from $ p_d $ and the samples generated by the model. As a result, the distribution defined by $ G $ (denoted as $ p_g $) gradually becomes closer to $ p_d $. Crucially, and unlike previous work, our training procedure does not require any strong supervision in the form of aligned examples from $ p_d $ and $ p_a $.

We give the concrete training objectives for $ G $ and $ D $ below.

\subsection{Objectives}

In our experiments, we found that the original minimax objective from \citet{Goodfellow14} was hard to optimize in the context of our model, so we opted for using a variant that employs Wasserstein distance as a measure of divergence between distributions \cite{Gulrajani17}, as it appears to handle vastly dissimilar $ p_g $ and $ p_d $ more gracefully. In this case, the discriminator is considered ``confused'' if it assigns the same scores (in expectation) to the inputs coming both from $ p_d $ and $ p_g $. We note that our approach can be used in conjunction with other forms of GAN objectives as well.

\textbf{Discriminator.} Following \cite{Gulrajani17}, we define the objective for $ D $ as:
\begin{equation}
    \mathcal{L}_{D} = -\mathbb{E}_{\xx \sim p_d} \left[ D(\xx) \right] + \mathbb{E}_{\xx \sim p_g} \left[ D(\xx) \right] + R,
\label{eq:discriminator_objective}
\end{equation}
where $ R $ is a regularization term softly constraining $ D $ to stay in the set of Lipschitz continuous functions (for some fixed Lipschitz constant). It is worth noting that the solution to \eq{discriminator_objective} is defined up to an additive constant. Due to the nature of how we use discriminator outputs during training of $ \pi $, we found it beneficial to resolve this ambiguity by encouraging $ D(\xx) $ to be close to $ 0 $ on average for $ \xx \sim \frac{1}{2} p_g + \frac{1}{2} p_d $.

\textbf{Generator.} We formally define $ \pi $ as a network that at every time step $ t $ predicts a distribution over all possible commands $ \pi_t = \pi(a_t | s_t; \theta) $, where $ s_t $ is the recurrent state of the network, and $ \theta $ is a set of learnable parameters. Given a sequence of samples $ \left( a_t \, | \, a_t \sim \pi_t, 1 \leq t \leq N \right) $, a sample from $ p_g $ is then computed as $ \mathcal{R}(a_1, a_2, \ldots, a_N) $. Since the generator is an arbitrary non-differentiable function, we cannot optimize
\begin{equation}
    \mathcal{L}_{G} = -\mathbb{E}_{\xx \sim p_g} \left[ D(\xx) \right]
\label{eq:generator_objective}
\end{equation}
with naive gradient descent. Therefore we pose this problem as maximization of the expected return which can be solved using standard techniques from reinforcement learning. Specifically, we employ a variant of the REINFORCE \cite{Williams92} algorithm, advantage actor-critic (A2C):
\begin{equation}
    \mathcal{L}_{G} = -\sum_t \log \pi(a_t \, | \, s_t; \theta) \left[ R_t - V^{\pi}(s_t) \right] \, ,
\label{eq:policy_gradients}
\end{equation}
where $ V^{\pi} $ is an approximation to the value function which is considered to be independent of $ \theta $, and $ R_t = \sum_t^N r_t $ is a 1-sample Monte-Carlo estimate of the return. Optimizing \eq{policy_gradients} recovers the solution to \eq{generator_objective} if the rewards are set to: 
\begin{equation}
    r_t = 
    \begin{cases}
        0 \, , & t < N \, , \\
        D(\mathcal{R}(a_1, a_2, \ldots, a_N)) \, , & t = N \, .
    \end{cases}
\label{eq:rewards}
\end{equation}
One interesting aspect of this new formulation is that we can also bias the search by introducing intermediate rewards which may depend not only on the output of $ \mathcal{R} $ but also on commands used to generate that output. We present several examples of such rewards in \sect{experiments}.

\subsection{Conditional Generation}
\label{sect:conditional_generation}

So far, we have described the case of unconditional generation, but in many situations it is useful to condition the model on auxiliary input \cite{Mirza14}. For instance, one might be interested in finding a specific program that generates a given image $ \xxsub{target} $. That could be achieved by supplying $ \xxsub{target} $ both to the policy and to the discriminator networks. In other words, 
\begin{equation}
    p_g = \mathcal{R} \left( p_a(a | \xxsub{target}) \right) \, ,
\end{equation}
while $ p_d $ becomes a Dirac $ \delta $-function centered at $ \xxsub{target} $. The first two terms in \eq{discriminator_objective} thus reduce to
\begin{equation}
    -D(\xxsub{target} \, | \, \xxsub{target}) + \mathbb{E}_{\xx \sim p_g} \left[ D(\xx | \xxsub{target}) \right] \, .
\label{eq:discriminator_wgan}
\end{equation}
It can be shown that for this particular setting of $ p_g $ and $ p_d $, the $ \ell^2 $-distance is an optimal discriminator. However, in general it is not a unique solution to \eq{discriminator_objective} and may be a poor candidate to be used as the generator's reward signal (see \sect{optimal_d} in the appendix for details). In \sect{experiments}, we empirically evaluate both $ \ell^2 $ and a dynamically learned $ D $ and conclude that those two options are not equivalent in practice (for example, see \fig{disc_vs_l2_mnist_omniglot}).

\subsection{Distributed Learning}

Our training pipeline is outlined in \fig{distributed_training}. It is an extension of the recently proposed \textit{IMPALA} architecture \cite{Espeholt18}. For training, we define three kinds of workers:
\begin{itemize}
    \item \emph{Actors} are responsible for generating the training trajectories through interaction between the policy network and the rendering simulator. Each trajectory contains a sequence $ ((\pi_t, a_t) \, | \, 1 \leq t \leq N) $ as well as all intermediate renderings produced by $ \mathcal{R} $.
    \item A \emph{policy learner} receives trajectories from the actors, combines them into a batch and updates $ \pi $ by performing an SGD step on $\mathcal{L}_G$ \eq{generator_objective}. Following common practice \cite{Mnih16}, we augment $ \mathcal{L}_G $ with an entropy penalty encouraging exploration.
    \item In contrast to the base IMPALA setup, we define an additional \emph{discriminator learner}.
    This worker consumes random examples from $ p_d $, as well as generated data (final renders) coming from the actor workers, and optimizes $ \mathcal{L}_D $ \eq{discriminator_objective}.
\end{itemize}

In the original paper introducing WGAN with gradient penalty \cite{Gulrajani17}, the authors note that in order to obtain better performance, the discriminator has to be updated more frequently than the generator. In our setting, generation of each model sample is expensive since it involves multiple invocations of an external simulator. We therefore do not omit any trajectories in the policy learner. Instead, we decouple the $ D $ updates from the $ \pi $ updates by introducing a \emph{replay buffer} that serves as a communication layer between the actors and the discriminator learner.
That allows the latter to optimize $ D $ at a higher rate than the training of the policy network due to the difference in network sizes ($ \pi $ is a multi-step RNN, while $ D $ is a plain CNN). We note that even though sampling from a replay buffer inevitably leads to smoothing of $ p_g $, we found this setup to work well in practice.

\section{Experiments}
\label{sect:experiments}

\subsection{Datasets}

We validate our approach on three real-world and one synthetic image dataset. The first, \dataset{MNIST} \cite{Lecun98}, is regarded as a standard sanity check for newly proposed generative models. It contains \num{70000} examples of handwritten digits, of which \num{10000} constitute a test set. Each example is a $ 28 \times 28 $ grayscale image. Although the dataset is often considered ``solved'' by neural decoder-based approaches (including GANs and VAEs), these approaches do not focus on recovering interpretable structure from the data. We, therefore, choose not to discard \dataset{MNIST} from the empirical evaluation since it is likely that additional constraints increase the difficulty of the modeling task.

The second dataset, \dataset{Omniglot} \cite{Lake15}, comprises \num{1623} handwritten characters from 50 alphabets. Compared to \dataset{MNIST}, this dataset introduces three additional challenges: higher data variability, higher complexity of symbols (\eg, disjoint subcurves) and fewer (only 20) data points per symbol class.

Since both \dataset{MNIST} and \dataset{Omniglot} represent a restricted line drawing domain, we diversify our set of experiments by testing the proposed method on \dataset{CelebA} \cite{Liu15}. The dataset contains over \num{200000} color headshots of celebrities with large variation in poses, backgrounds and lighting conditions.

Lastly, we are interested in evaluating our approach on the task of unsupervised 3D scene understanding which is a crucial precursor for manipulating and reasoning about objects in the real world. To that end, we created a procedural dataset called \dataset{MuJoCo Scenes} consisting of renders of simple 3D primitives (up to 5 objects) scattered around a square platform (see \fig{mujoco_reconstruction}). The training set is comprised of \num{50000} RGB images generated by means of the \texttt{MuJoCo} environment discussed in the next section. 

In each case, we rescale the images to $ 64 \times 64 $, which allows us to reuse the same network architectures in all the experiments, demonstrating the generality of our method.

\subsection{Environments}
\label{sect:environments}

\begin{figure}[t!]
\centering
\colorlet{linecolor}{black}
\definecolor{strokegreen}{HTML}{4AA54A}
\begin{tikzpicture}[
  every text node part/.style={align=right},
  linecolor, text=black,
  node distance=4mm,
  point/.style={
    anchor=center,
    circle,minimum size=0.25cm,
    inner sep=0pt,
    line width=1.5pt,draw=linecolor,fill=none,
    font=\huge}]

\node (strokes) [anchor=south west,inner sep=0] {\pgfimage[width=0.34\textwidth]{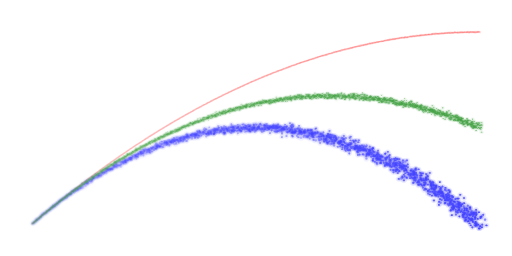}};
\begin{scope}[x={(strokes.south east)},y={(strokes.north west)}]
    \node (current) [point] at (4/64,4/32) {};
    \node (control) [point] at (32/64,28/32) {};
    \node (end) [point] at (60/64,16/32) {};
    \node (aux_end_1) [point,opacity=0.25] at (60/64,4/32) {};
    \node (aux_end_2) [point,opacity=0.25] at (60/64,28/32) {};
\end{scope}

\node[right=2pt of current,font=\scriptsize] {Current location, $ l_t $};
\node[left=2pt of control,font=\scriptsize] {Control point, $ a_t^1 = p_c $};
\node[right=0pt of end,font=\scriptsize] {\setlength{\tabcolsep}{2pt}\renewcommand{\arraystretch}{1.3}\begin{tabular}{l r}%
    End point, & $ a_t^2 = l_{t+1} $\\
    \textbf{Pressure}, & $ a_t^3 $\\
    Brush size, & $ a_t^4 $\\
    \textcolor{strokegreen}{Color}, & $ a_t^{\{5,6,7\}} $\\
\end{tabular}};

\draw[line width=1.5pt,draw=black!80,opacity=0.8] (current) -- (control);
\draw[line width=1.5pt,draw=black!80,opacity=0.8] (control) -- (end);
\draw[line width=1.5pt,draw=black!80,opacity=0.3] (control) -- (aux_end_1);
\draw[line width=1.5pt,draw=black!80,opacity=0.3] (control) -- (aux_end_2);

\end{tikzpicture}%
\caption{Illustration of the \textbf{agent's action space} in the \texttt{libmypaint} environment. We show three different strokes (red, green, blue) that can result from a single instruction from the agent to the renderer. Starting from a position on the canvas, the agent selects the coordinates of the next end point, the coordinates of the intermediate control point, as well as the brush size, pressure and color. See \sect{environments} for details.\vspace{-1mm}}
\label{fig:libmypaint_action_space}
\end{figure}
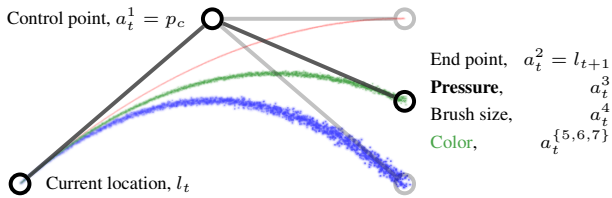

We introduce two new rendering environments. For \dataset{MNIST}, \dataset{Omniglot} and \dataset{CelebA} generation we use an open-source painting library \texttt{libmypaint} \cite{Libmypaint}. The agent controls a brush and produces a sequence of (possibly disjoint) strokes on a canvas $ C $. The state of the environment is comprised of the contents of $ C $ as well as the current brush location $ l_t $. Each action $ a_t $ is a tuple of 8 discrete decisions $ (a_t^1, a_t^2, \ldots, a_t^8) $ (see \fig{libmypaint_action_space}). The first two components are the control point $ p_c $ and the end point $ l_{t + 1} $ of the stroke, which is specified as a quadratic B\'{e}zier curve:
\begin{equation}
    p(\tau) = (1 - \tau)^2 \, l_t + 2 (1 - \tau) \, \tau \, p_c + \tau^2 \, l_{t + 1} \, ,
\end{equation}
where $ \tau \in [0, 1] $. In our experiments, we define the valid range of locations as a $ 32 \times 32 $ grid imposed on $ C $. We set $ l_0 $ to the upper left corner of the canvas. The next 5 components represent the appearance of the stroke: the pressure that the agent applies to the brush (10 levels), the brush size, and the stroke color characterized by mixture of red, green and blue (20 bins for each color component). The last element of $ a_t $ is a binary flag specifying the type of action: the agent can choose either to produce a stroke or to jump right to $ l_{t + 1} $. For grayscale datasets (\dataset{MNIST} and \dataset{Omniglot}), we omit the color components.

In the \dataset{MuJoCo Scenes} experiment, we render images using a \texttt{MuJoCo}-based environment \cite{Todorov12}. At each time step, the agent has to decide on the object type (4 options), its location on a $ 16 \times 16 $ grid, its size (3 options) and the color (3 color components with 4 bins each). The resulting tuple is sent to the environment, which adds an object to the scene according to the specification. Additionally, the agent can decide to skip a move or change the most recently emitted object. All three types of actions are illustrated in \fig{mujoco_actor_diagram}.

\subsection{\dataset{MNIST}}

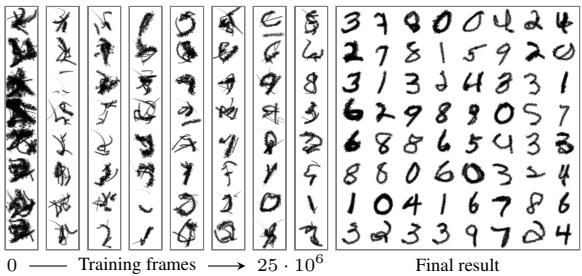
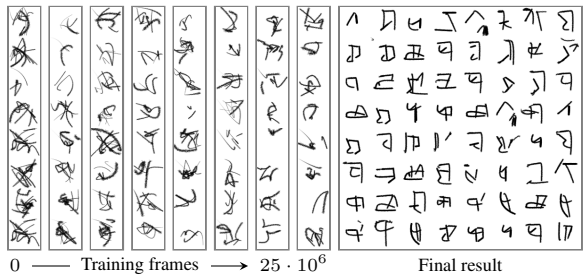
\begin{figure}[t!]
    \centering
    \begin{subfigure}[b]{0.48\textwidth}
        \centering
        \begin{tikzpicture}
    \foreach \x/\filename[evaluate=\x as \xmo using {int(\x-1)}] in {0/1280,1/3847680,2/7658240,3/10720000,4/14536960,5/18346240,6/21930240,7/25137920,8/1419539200} {
        \node [inner sep=0,anchor=north west,draw=black!50,ultra thick] (\x) at (\x * 0.55cm, 0) {\pgfimage[height=3.2cm]{./figures/experiments/mnist/generation/\filename.png}};
    }
    \node[fit=(0)(7),inner sep=0] (progression) {};
    \node[inner sep=0,anchor=north west,below right=0.05cm] (left_bound) at (progression.south west) {\scriptsize $ 0 $\vphantom{$ 10^6 $}};
    \node[inner sep=0,anchor=north east,below left=0.05cm] (right_bound) at (progression.south east) {\scriptsize $ 25 \cdot 10^6 $};
    \draw[yshift=-1cm,shorten >=0.15cm,shorten <=0.15cm,-stealth] ($ (left_bound.south east) + (0,2pt) $) -- ($ (right_bound.south west) + (0,2pt) $) node[midway,inner xsep=0.15cm,inner ysep=0,fill=white] {\scriptsize Training frames};
    \node[inner sep=0,anchor=north,below=0.05cm] (final) at (8.south) {\scriptsize Final result\vphantom{$ 10^6 $}};
\end{tikzpicture}
        \caption{MNIST unconditional generation}
        \label{fig:mnist_unconditional_generation}
    \end{subfigure} \\
    \vspace{0.4cm}
    \begin{subfigure}[b]{0.48\textwidth}
        \centering
        \begin{tikzpicture}
    \foreach \y in {0,...,7} {
        \foreach \x[evaluate=\x as \filename using {int(\y*8+\x)}] in {0,...,7} {
            \node [inner sep=0,anchor=north west,draw=black!50,ultra thick] (\x) at (\x * 0.95cm, \y * 0.55cm) {\pgfimage[height=0.4cm]{./figures/experiments/mnist/reconstruction/\filename.png}};
        }
    }
\end{tikzpicture}
        \caption{MNIST reconstruction}
        \label{fig:mnist_reconstruction}
    \end{subfigure} %
    \caption{\textbf{MNIST.} \textbf{(a)} A SPIRAL agent is trained to draw MNIST digits via a sequence of strokes in the \texttt{libmypaint} environment. As training progresses, the quality of the generations increases. The final samples capture the multi-modality of the dataset, varying brush sizes and digit styles. \textbf{(b)} A conditional SPIRAL agent is trained to reconstruct using the same action space. Reconstructions (left) match ground-truth (right) accurately.\vspace{-2mm}}
\end{figure}

For the \dataset{MNIST} dataset, we conduct two sets of experiments. In the first set, we train an unconditional agent to model the data distribution. Along with the reward provided by the discriminator we also use auxiliary penalties expressing our inductive biases for the particular type of data. To encourage the agent to draw a digit in a single continuous motion of the brush, we provide a small negative reward for starting each continuous sequence of strokes. We also found it beneficial to penalize our model for not producing any visible strokes at all. The resulting agent manages to generate samples clearly recognizable as hand-written digits. Examples of such generations are shown in \fig{mnist_unconditional_generation}.

In the second set of experiments, we train an agent to generate the strokes for a given target digit, and we compare two kinds of rewards discussed in \sect{conditional_generation}: fixed $ \ell^2 $-distance and the discriminator score. The results are summarized in \fig{disc_vs_l2_mnist_omniglot} (blue curves). We note that the discriminator-based approach significantly speeds up training of the model and achieves lower final $ \ell^2 $ error. When no auxiliary rewards were employed, $ \ell^2 $-based runs failed to learn reasonable reconstructions. \fig{mnist_reconstruction} presents several conditional generations produced by our method.

Following \cite{Sharma17}, we also train a ``blind'' version of the agent, \ie, we do not feed intermediate canvas states as an input to $ \pi $. That means that the model cannot rely on reactive behaviour since it does not ``see'' the immediate consequences of its decisions. The training curve for this experiment is shown in \fig{disc_vs_l2_mnist_omniglot} (dotted blue line). Although the agent does not reach the level of performance of the full model, it can still produce sensible reconstructions which suggests that our approach could be used in the more general setting of program synthesis, where access to intermediate states of the execution pipeline is not assumed.

\subsection{\dataset{Omniglot}}

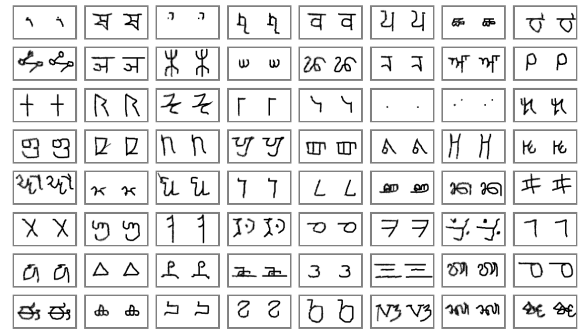
\begin{figure}[t!]
    \centering
    \begin{subfigure}[b]{0.48\textwidth}
        \centering
        \begin{tikzpicture}
    \foreach \x/\filename[evaluate=\x as \xmo using {int(\x-1)}] in {0/1280,1/3927040,2/7880960,3/11060480,4/15046400,5/18204160,6/22926080,7/25313280,8/1585785600} {
        \node [inner sep=0,anchor=north west,draw=black!50,ultra thick] (\x) at (\x * 0.55cm, 0) {\pgfimage[height=3.2cm]{./figures/experiments/omniglot/generation/\filename.png}};
    }
    \node[fit=(0)(7),inner sep=0] (progression) {};
    \node[inner sep=0,anchor=north west,below right=0.05cm] (left_bound) at (progression.south west) {\scriptsize $ 0 $\vphantom{$ 10^6 $}};
    \node[inner sep=0,anchor=north east,below left=0.05cm] (right_bound) at (progression.south east) {\scriptsize $ 25 \cdot 10^6 $};
    \draw[yshift=-1cm,shorten >=0.15cm,shorten <=0.15cm,-stealth] ($ (left_bound.south east) + (0,2pt) $) -- ($ (right_bound.south west) + (0,2pt) $) node[midway,inner xsep=0.15cm,inner ysep=0,fill=white] {\scriptsize Training frames};
    \node[inner sep=0,anchor=north,below=0.05cm] (final) at (8.south) {\scriptsize Final result\vphantom{$ 10^6 $}};
\end{tikzpicture}
        \caption{Omniglot unconditional generation}
        \label{fig:omniglot_unconditional generation}
    \end{subfigure} \\
    \vspace{0.4cm}
    \begin{subfigure}[b]{0.48\textwidth}
        \centering
        \begin{tikzpicture}
    \foreach \y in {0,...,7} {
        \foreach \x[evaluate=\x as \filename using {int(\y*8+\x)}] in {0,...,7} {
            \node [inner sep=0,anchor=north west,draw=black!50,ultra thick] (\x) at (\x * 0.95cm, \y * 0.55cm) {\pgfimage[height=0.4cm]{./figures/experiments/omniglot/reconstruction/\filename.png}};
        }
    }
\end{tikzpicture}
        \caption{Omniglot reconstruction}
        \label{fig:omniglot_reconstruction}
    \end{subfigure} %
    \caption{\textbf{Omniglot.} \textbf{(a)} A SPIRAL agent is trained to draw MNIST digits via a sequence of strokes in the \texttt{libmypaint} environment. As training progresses, the quality of the generations increase. The final samples capture the multi-modality of the dataset, varying brush sizes and character styles. \textbf{(b)} A conditional SPIRAL agent is trained to reconstruct using the same action space. Reconstructions (left) match ground-truth (right) accurately.\vspace{-2mm}}
\end{figure}

In the previous section, we showed that our approach works reasonably well for handwritten digits. In this series of experiments, we test our agent in a similar but more challenging setting of handwritten characters. The difficulty of the dataset manifests itself in lower quality of unconditional generations (\fig{omniglot_unconditional generation}). Note that this task appears to be hard for other neural network based approaches as well: models that do produce good samples, such as \cite{Rezende16}, do not do so in a manner that mimics actual strokes. 

The conditional agent, on the other hand, managed to reach convincing quality of reconstructions (\fig{omniglot_reconstruction}). Unfortunately, we could not make the $ \ell^2 $-based model work well in this setting (\fig{disc_vs_l2_mnist_omniglot}; dashed red line). This suggests not only that discriminator rewards speed up learning, but also that they allow successful training of agents in cases where na{\"i}ve rewards like $ \ell^2 $ do not result in sufficient exploration.

\begin{figure}[t!]
    \centering
    \begin{sc}
    \newcommand{\modeltestrow}[1]{%
\includegraphics[width=0.07\textwidth]{figures/experiments/omniglot/model_test/#1/input.png} &
\includegraphics[width=0.07\textwidth]{figures/experiments/omniglot/model_test/#1/lowres.png} &
\includegraphics[width=0.07\textwidth]{figures/experiments/omniglot/model_test/#1/hires.png} &
\includegraphics[width=0.07\textwidth]{figures/experiments/omniglot/model_test/#1/avg_sample.png}
}
\renewcommand{\arraystretch}{0.85}
\begin{tabular}{ c c c c }
 {\scriptsize Input} & {\scriptsize Reconstruction} & {\scriptsize Reconstruction} & {\scriptsize Trace} \\
 $ 64 \times 64 $ & $ 64 \times 64 $ & $ 256 \times 256 $ & $ 256 \times 256 $ \\
 \hline
 \rule{0pt}{1.3cm} \modeltestrow{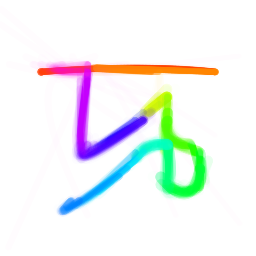} \\
 \modeltestrow{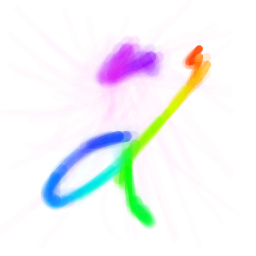} \\
 \modeltestrow{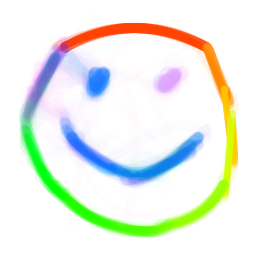}
\end{tabular}%
    \end{sc}
    \caption{\textbf{Image parsing} using the SPIRAL agent trained on Omniglot. All images from test sets. Given a rastered input (a), the agent produces a reconstruction (b) that closely matches the input. (c) Having access to the underlying strokes, we can render the character at a higher resolution, or in a different stroke style. (d) The agent effectively parses the input image into strokes. Each stroke is depicted in a separate color (we show average across 100 samples).\vspace{-2.5mm}}
    \label{fig:omniglot_model_test}
\end{figure}

Since \dataset{Omniglot} contains a highly diverse set of symbols, over the course of training our model could learn a general notion of image reproduction rather than simply memorizing dataset-specific strokes. In order to test this, we feed a trained agent with previously unseen line drawings. The resulting reconstructions are shown in \fig{omniglot_model_test}. The agent handles out-of-domain images well, although it is slightly better at reconstructing the \dataset{Omniglot} test set.

\subsection{\dataset{CelebA}}

Since the \texttt{libmypaint} environment is also capable of producing complex color paintings, we explore this direction by training a conditional agent on the \dataset{CelebA} dataset. As in previous experiments, we use 20-step episodes, and as before, the agent does not receive any intermediate rewards. In addition to the reconstruction reward (either $ \ell^2 $ or discriminator-based), we put a penalty on the earth mover's distance between the color histograms of the model's output and $ \xxsub{target} $. We found this relatively task-agnostic penalty to slightly improve the performance of the method, but we would like to stress that it is by no means necessary.

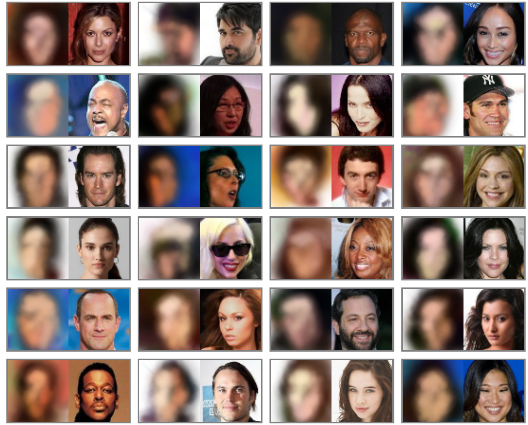
\begin{figure}[t!]
    \centering
    \begin{tikzpicture}
    \foreach \y in {0,...,5} {
        \foreach \x[evaluate=\x as \filename using {int(\y*4+\x)}] in {0,...,3} {
            \node [inner sep=0,anchor=north west,draw=black!50,ultra thick] (\x) at (\x * 1.75cm, \y * 0.95cm) {\pgfimage[height=0.8cm]{./figures/experiments/celeba/reconstruction2/\filename.png}};
        }
    }
\end{tikzpicture}
    \caption{\textbf{\dataset{CelebA} reconstructions.} The SPIRAL agent reconstructs human faces in 20 strokes. Although blurry, the reconstructions closely match the high-level structure of each image, for instance the background color, the position of the face and the color of the person's hair. In some cases, shadows around eyes and the nose are visible.\vspace{-2.5mm}}
    \label{fig:celeba_reconstruction}
\end{figure}

Given that we made no effort whatsoever to adapt the action space for this domain, it is not surprising that it takes significantly more time to discover the policy that produces images resembling the target (\fig{celeba_reconstruction}). As in the \dataset{Omniglot} experiment, the $ \ell^2 $-based agent demonstrates some improvement over the random policy but gets stuck and, as a result, fails to learn sensible reconstructions (\fig{disc_vs_l2_celeba_mujoco}).

Although blurry, the model's reconstruction closely matches the high-level structure of each image. For instance the background color, the position of the face and the color of the person's hair. In some cases, shadows around eyes and the nose are visible. However, we observe that our model tends to generate strokes in the first half of the episode that are fully occluded by strokes in the second half. We hypothesize that this phenomenon is a consequence of credit assignment being quite challenging in this task. One possible remedy is to provide the agent with a mid-episode reward for reproducing a blurred version of the target image. We leave this prospect for future work.

\subsection{\dataset{MuJoCo Scenes}}

For the \dataset{MuJoCo Scenes} dataset, we use our agent to construct simple CAD programs that best explain input images. Here we are only considering the case of conditional generation. Like before, the reward function for the generator can be either the $ \ell^2 $ score or the discriminator output. We did not provide any auxiliary reward signals. The model is unrolled for 20 time steps, so it has the capacity to infer and represent up to 20 objects and their attributes. As we mentioned in \sect{environments}, the training data consists of scenes with at most 5 objects. The agent does not have this knowledge a priori and needs to learn to place the right number of primitives.

\begin{figure}[ht!]
    \centering
    \newcounter{mujocoexampleidx}\setcounter{mujocoexampleidx}{1}
\begin{tikzpicture}
    \foreach \y in {0,...,5} {
        \foreach \x[evaluate=\x as \filename using {int(\y*8+\x)}] in {0,...,3} {
            \node [inner sep=0,anchor=north west,draw=black!50,ultra thick] (\x) at (\x * 1.75cm, \y * 0.95cm) {\pgfimage[height=0.8cm]{./figures/experiments/mujoco/reconstruction/ren\themujocoexampleidx.png}\pgfimage[height=0.8cm]{./figures/experiments/mujoco/reconstruction/target\themujocoexampleidx.png}};
            \stepcounter{mujocoexampleidx}
        }
    }
\end{tikzpicture}
    \caption{\textbf{3D scene reconstructions.} The SPIRAL agent is trained to reconstruct 3D scenes by emitting sequences of commands for the \texttt{MuJoCo} environment. In each pair, the left image corresponds to the model's output while the right one is the target. Our method is capable of acurately inferring the number of objects, their locations, sizes and colors.\vspace{-1mm}}
    \label{fig:mujoco_reconstruction}
\end{figure}
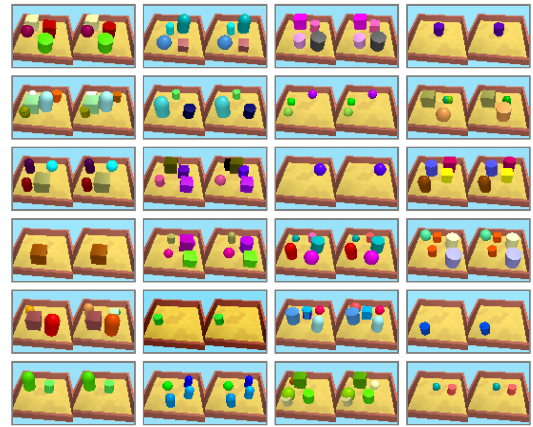

As shown in \fig{disc_vs_l2_celeba_mujoco}, the agent trained to directly minimize $ \ell^2 $ is unable to solve the task and has significantly higher pixel-wise error. In comparison, the discriminator-based variant solves the task and produces near-perfect reconstructions on a holdout set (\fig{mujoco_reconstruction}).

We note that our agent has to deal with a high-cardinality action space intractable for a brute-force search. Indeed, the total number of possible execution traces is $ M^N $, where $ M = 4 \cdot 16^2 \cdot 3 \cdot 4^3 \cdot 3 $ is the total number of attribute settings for a single object (see \sect{environments} for details) and $ N = 20 $ is the length of an episode.\footnote{The actual number of scene configurations is smaller but still intractable.} In order to demonstrate the computational hardness of the task, we ran a general-purpose \emph{Metropolis-Hastings} inference algorithm on a set of 100 images. The algorithm samples an execution trace defining attributes for a maximum of 20 primitives. These attributes are treated as latent variables. During each time step of inference, a block of attributes (including the presence/absence flag) corresponding to a single object is flipped uniformly within appropriate ranges. The resulting trace is rendered by the environement into an output sample which is then accepted or rejected using the Metropolis-Hastings update rule, with a  Gaussian likelihood centered around the test image and a fixed diagonal covariance of $ 0.25 $. As shown in \fig{spiral_vs_mcmc}, the MCMC search baseline was unable to solve the task even after a large number of evaluations.

\tikzsetexternalprefix{figures/experiments/plots/}
\tikzexternalenable
\ifshowplots

\begin{figure*}[t!]
    \centering
    \begin{subfigure}[b]{0.48\textwidth}
        \centering
        \setlength\figureheight{6cm}%
        \setlength\figurewidth{8cm}%
        \tikzsetnextfilename{disc_vs_l2_mnist_omniglot}%
\begin{tikzpicture}

\end{tikzpicture}
        \caption{\dataset{MNIST} and \dataset{Omniglot}}
        \label{fig:disc_vs_l2_mnist_omniglot}
    \end{subfigure} %
    ~
    \begin{subfigure}[b]{0.48\textwidth}
        \centering
        \setlength\figureheight{6cm}%
        \setlength\figurewidth{8cm}%
        \tikzsetnextfilename{disc_vs_l2_celeba_mujoco}%
\begin{tikzpicture}

\end{tikzpicture}
        \caption{\dataset{CelebA} and \dataset{MuJoCo Scenes}}
        \label{fig:disc_vs_l2_celeba_mujoco}
    \end{subfigure} %
    \vspace{-1mm} %
\caption{\textbf{$ \ell^2 $-distance between reconstructions and ground truth} images over the course of training. Across all datasets, we observe that training using a discriminator leads to significantly lower $ \ell^2 $-distances, than when directly minimizing $ \ell^2 $. We also show in (a) that the SPIRAL agent is capable of reconstructing even when it does not have access to the renderer in intermediate steps, however this does lead to a small degradation in performance.\vspace{-1mm}}
\label{fig:disc_vs_l2_plot}
\end{figure*}
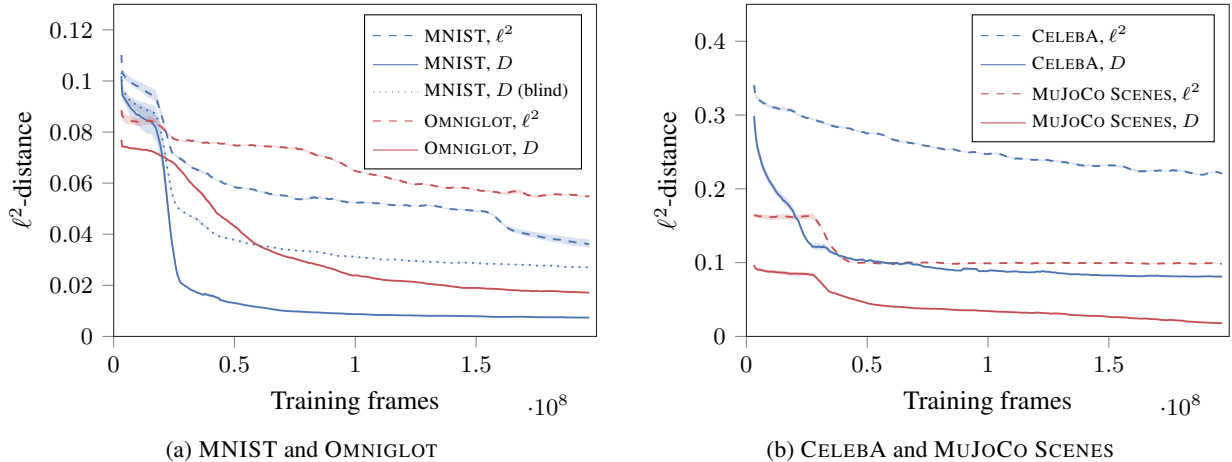

\fi
\tikzexternaldisable

\tikzexternalenable
\ifshowplots

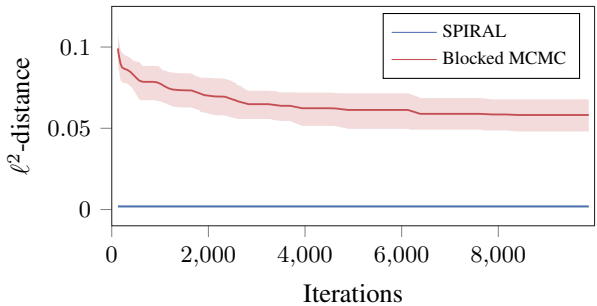
\begin{figure}[h!]
\centering
\setlength\figureheight{4.5cm}%
\setlength\figurewidth{8cm}%
\tikzsetnextfilename{spiral_vs_mcmc}%
\begin{tikzpicture}

\end{tikzpicture} %
\vspace{-1mm} %
\caption{\textbf{Blocked Metropolis-Hastings (MCMC) \textit{vs} SPIRAL.} The \dataset{MuJoCo Scenes} dataset has a large combinatorial search space. We ran a general-purpose MCMC algorithm with object based blocked proposals and SPIRAL on 100 holdout images during inference time. SPIRAL reliably processes every image in a single pass. We ran the MCMC algorithm for thousands of evaluations but it was unable to solve the task.\vspace{-2.5mm}}
\label{fig:spiral_vs_mcmc}
\end{figure}

\fi
\tikzexternaldisable
\section{Discussion}

Scaling visual program synthesis to real world and combinatorial datasets has been a challenge. We have shown that it is possible to train an adversarial generative agent employing black-box rendering simulators. Our results indicate that using the Wasserstein discriminator's output as a reward function with asynchronous reinforcement learning can provide a scaling path for visual program synthesis. The current exploration strategy used in the agent is entropy-based, but future work should address this limitation by employing sophisticated search algorithms for policy improvement. For instance, Monte Carlo Tree Search can be used, analogous to AlphaGo Zero \cite{Silver17}. General-purpose inference algorithms could also be used for this purpose. 

Future work should explore different parameterizations of action spaces. For instance, the use of two arbitrary control points is perhaps not the best way to represent strokes, as it is hard to deal with straight lines. Actions could also directly parametrize 3D surfaces, planes and learned texture models to invert richer visual scenes. On the reward side, using a joint image-action discriminator similar to BiGAN/ALI \cite{Donahue16,Dumoulin16} (in this case, the policy can viewed as an encoder, while the renderer becomes a decoder) could result in a more meaningful learning signal, since $ D $ will be forced to focus on the semantics of the image.

We hope that this paper provides an avenue to further explore inverse simulation and program synthesis on applications ranging from vision, graphics, speech, music and scientific simulators.

\clearpage

\section*{Acknowledgements}

We would like to thank Mike Chrzanowski, Michael Figurnov, David Warde-Farley, Pushmeet Kohli, Sasha Vezhnevets and Suman Ravuri for helping with the manuscript preparation as well as Sergey Bartunov, Ian Goodfellow, Jacob Menick, Lasse Espeholt, Ivo Danihelka, Junyoung Chung, \c{C}a\u{g}lar G\"{u}l\c{c}ehre, Dzmitry Bahdanau and Koray Kavukcuoglu for insightful discussions and support.

\bibliography{references}
\bibliographystyle{icml2018}

\clearpage

\appendix
\section{Optimal $ D $ for Conditional Generation}
\label{sect:optimal_d}

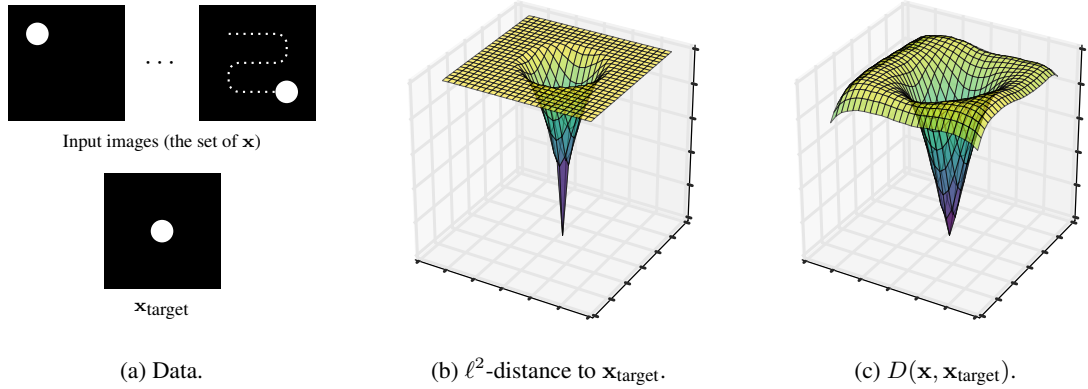
\begin{figure*}[t!]
    \centering
    \begin{subfigure}[b]{0.3\textwidth}
        \centering
        \begin{tikzpicture}
            \node (placeholder) [minimum size=0.95\textwidth,inner sep=0,outer sep=0] {};
            \node (image_1) [anchor=north west,inner sep=0,outer sep=0,minimum size=0.3\textwidth,fill=black] at ($ (placeholder.north west) + (0.4cm,-0.4cm) $) {};
            \begin{scope}[shift={(image_1.north west)},x={(image_1.north east)},y={(image_1.south west)}]
                \node [circle,inner sep=0,minimum size=0.3cm,fill=white] at (0.25,0.25) {};
            \end{scope}
            \node (image_2) [anchor=north east,inner sep=0,outer sep=0,minimum size=0.3\textwidth,fill=black] at ($ (placeholder.north east) + (-0.4cm,-0.4cm) $) {};
            \begin{scope}[shift={(image_2.north west)},x={(image_2.north east)},y={(image_2.south west)}]
                \node [circle,inner sep=0,minimum size=0.3cm,fill=white] at (0.75,0.75) {};
                \draw [draw=white,rounded corners,thick,dotted] (0.25,0.25) -- (0.75,0.25) -- (0.75,0.5) -- (0.25,0.5) -- (0.25,0.75) -- (0.75,0.75);
            \end{scope}
            \node at ($ (image_1)!0.5!(image_2) $) {$ \cdots $};
            \node (top_images_bb) [fit=(image_1)(image_2),inner sep=0] {};
            \node [below=0.17cm of top_images_bb,inner sep=0,font=\scriptsize] {Input images (the set of $ \xx $)};
            \node (target_label) [anchor=south,inner sep=0,font=\scriptsize] at ($ (placeholder.south) + (0,0.3cm) $) {$ \xxsub{target} $};
            \node (image_3) [anchor=south,inner sep=0,outer sep=0,minimum size=0.3\textwidth,fill=black] at ($ (target_label.north) + (0cm,0.17cm) $) {};
            \begin{scope}[shift={(image_3.north west)},x={(image_3.north east)},y={(image_3.south west)}]
                \node [circle,inner sep=0,minimum size=0.3cm,fill=white] at (0.5,0.5) {};
            \end{scope}
        \end{tikzpicture}
        \caption{Data.\vphantom{$ \ell^2 $}}
    \end{subfigure}%
    \begin{subfigure}[b]{0.3\textwidth}
        \centering
        \begin{tikzpicture}
            \node (placeholder) [minimum size=0.95\textwidth,inner sep=0,outer sep=0] {};
            \node [anchor=center,inner sep=0,outer sep=0] at (placeholder.center) {\includegraphics[width=0.94\textwidth]{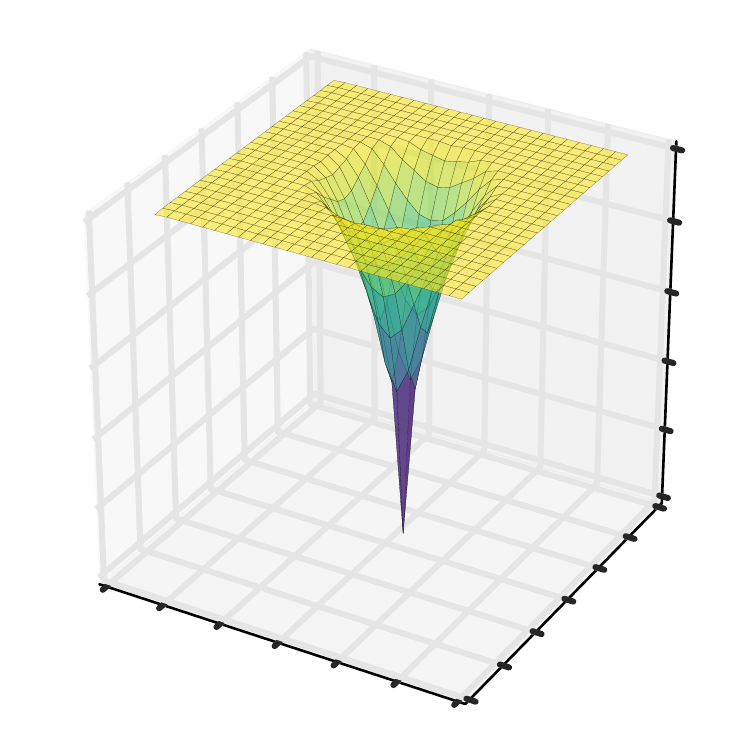}};
        \end{tikzpicture}
        \caption{$ \ell^2 $-distance to $ \xxsub{target} $.}
    \end{subfigure}%
    \begin{subfigure}[b]{0.3\textwidth}
        \centering
        \begin{tikzpicture}
            \node (placeholder) [minimum size=0.95\textwidth,inner sep=0,outer sep=0] {};
            \node [anchor=center,inner sep=0,outer sep=0] at (placeholder.center) {\includegraphics[width=0.94\textwidth]{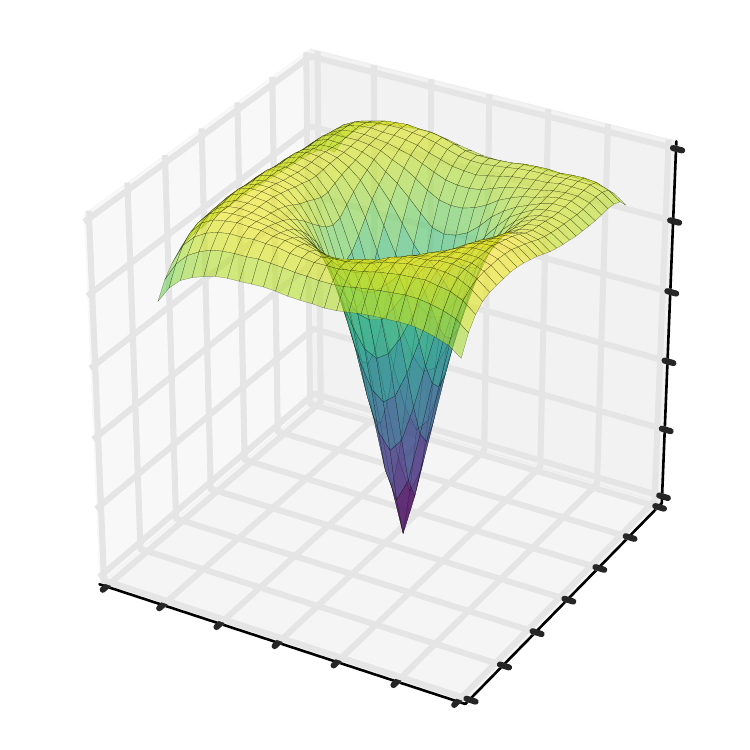}};
        \end{tikzpicture}
        \caption{$ D(\xx, \xxsub{target}) $.\vphantom{$ \ell^2 $}}
    \end{subfigure}%
    \caption{A toy experiment illustrating the \textbf{difference between $ \ell^2 $ and discriminator} training in practice. \textbf{(a)} We collect a dataset of images with a single solid circle in all possible locations (top) and pick one of them as a target image (bottom). \textbf{(b)} The $ \ell_2 $-distance (in the pixel space) from the input images to the target as a function of the circle location; the surface is flat around the borders since the circles do not overlap. \textbf{(c)} We train a discriminative model $ D $ that takes a pair of images and tells whether they match or not; just like $ \ell^2 $, the resulting function has a pit centered at the target location, but unlike (b), the surface now has better behaved gradients.}
    \label{fig:distance_surfaces}
\end{figure*}

It turns out that in the case of conditional generation (\ie, $ p_d $ is a Dirac $ \delta $-function), we can derive an explicit form of the optimal (non-parametric) discriminator $ D $. Indeed, \eq{discriminator_objective} corresponds to the dual representation of the Wasserstein-1 metric \cite{Villani08}. The primal form of that metric is defined as
\begin{equation}
    W_1(p_g, p_d) = \inf_{\gamma \in \Gamma(p_g, p_d)} \int \norm{\xx - \yy}_2 \, d \gamma(\xx, \yy) \, ,
\label{eq:wasserstein_primal}
\end{equation}
where $ \Gamma(p_g, p_d) $ is a set of all couplings between $ p_g $ and $ p_d $. Taking into account that the data distribution is a point mass, we can simplify \eq{wasserstein_primal}:
\begin{equation}
    W_1(p_g, p_d) = \mathbb{E}_{\xx \sim p_g} \norm{\xx - \xxsub{target}}_2 \, .
\end{equation}
The expression above gives the optimal value for $ \mathcal{L}_D $ in \eq{discriminator_objective}. Therefore $ D(\xx) = \norm{\xx - \xxsub{target}}_2 $ is a solution of \eq{discriminator_objective}:
\begin{equation}
\begin{split}
    \mathcal{L}_D (\norm{\xx - \xxsub{target}}_2) = &-\norm{\xxsub{target} - \xxsub{target}}_2 \\
                                                    &+ \mathbb{E}_{\xx \sim p_g} \norm{\xx - \xxsub{target}}_2 \\
                                                    &+ 0 \, ,
\end{split}
\end{equation}
where the last term ($ R $) is zero since the Euclidean distance belongs to the set of 1-Lipschitz functions.

This result suggests that for inverse graphics, in \eq{rewards}, one may use a fixed image distance (like the Euclidean distance $ \ell^2 $) instead of a parametric function optimized via the WGAN objective. Note, however, that $ \ell^2 $ is not a unique solution to \eq{discriminator_objective}. Consider, for example, the case where the model distribution is also a Dirac delta centered at $ \xxsub{g} $. The Wasserstein distance is equal to $ d = \norm{\xxsub{g} - \xxsub{target}}_2 $. In order to achieve that value in \eq{discriminator_objective}, we could take any $ D $ such that $ \forall \, \alpha \in [0, 1] $
\begin{equation}
    D \left( \alpha \, \xxsub{g} + (1 - \alpha) \, \xxsub{target} \right) = \alpha \cdot d \, ,
\label{eq:d_constraint}
\end{equation}
and $ \Lip(D) \leq 1 $. One example of such function would be a hyperplane $ H $ containing the segment \eq{d_constraint}. Let us now consider a set of points
\begin{equation}
    V = \left\{ \xx \, | \norm{\xx - \xxsub{g}}_2 < \varepsilon, \norm{\xx - \xxsub{target}}_2 = d \right\}
\end{equation}
in an $ \varepsilon $-vicinity of $ \xxsub{g} $. By definition, $ \norm{\xx - \xxsub{target}}_2 $ is constant for any $ \xx \in V $. That means that $ \ell^2 $ expresses no preference over points that are equidistant from $ \xxsub{target} $ even though some of them may be semantically closer to $ \xxsub{target} $. This property may significantly slow down learning if we are relying on $ \ell^2 $ (or similar distance) as our training signal. Functions like $ H $, on the other hand, have non-zero slope in $ V $ and therefore can potentially shift the search towards more promising subspaces.

One other reason why discriminator training is different from using a fixed image distance is that in practice, we do not optimize the exact dual formulation of the Wasserstein distance and, on top of that, use stochastic gradient descent methods which we do not run until convergence. A toy example illustrating that difference is presented in \fig{distance_surfaces}.

\todo[Maybe mention natural curriculum]

\section{Network Architectures}

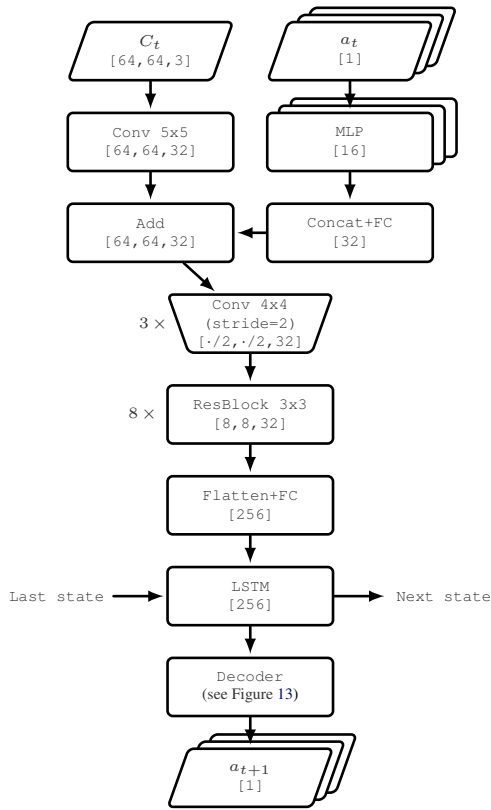
\begin{figure}
    \centering
    \begin{tikzpicture}[
  scale=0.3,
  black!100,
  text=black!80,
  node distance=0.4cm and 0.7cm,
  fnode/.style={
    align=center,
    rectangle,minimum height=22pt,minimum width=2.2cm,rounded corners=2pt,
    inner sep=3pt,
    line width=1pt,
    fill=pnodefill,draw=pnodedraw,
    font=\tiny\ttfamily},
  nonode/.style={
    align=center,
    font=\tiny\ttfamily},
  dnode/.style={
    fnode,trapezium,trapezium left angle=70, trapezium right angle=110,trapezium stretches},
  inode/.style={
    dnode,fill=inodefill,draw=inodedraw},
  downnode/.style={
    fnode,trapezium,trapezium left angle=110, trapezium right angle=110,trapezium stretches},
  pnode/.style={
    fnode,fill=pnodefill,draw=pnodedraw},
  pnode2/.style={
    fnode,fill=pnodefill!50!white,draw=pnodedraw!50!white},
  onode/.style={
    dnode,fill=onodefill,draw=onodedraw},
  smnode/.style={
    mnode,fill=mnodefill,draw=mnodedraw},
  flow/.style={
    -latex,shorten >=1pt,line width=1pt,line cap=round,rounded corners=2pt,draw=pnodedraw,draw==\#1},
  flow2/.style={
    flow,draw=pnodedraw!50!white}]
  \newcommand{\mytrap}[3]{%
    \node (#1) [anchor=center,#2] {\phantom{$ C_t $}\\\phantom{[64,64,1]}}; \node[anchor=center,align=center,font=\tiny\ttfamily] at (#1.center) {#3};
  }

  \mytrap{data}{dnode}{$ C_t $\\{[64,64,3]}};
  \node (preconv) [pnode, below= of data] {Conv 5x5\\{[64,64,32]}};
  \path (data) edge[flow] (preconv);

  \mytrap{prevaction}{dnode, double copy shadow={shadow xshift=1ex, shadow yshift=0.6ex},right= of data}{$ a_t $\\{[1]}};
  \node (embedaction) [pnode, double copy shadow={shadow xshift=1ex, shadow yshift=0.6ex}, below= of prevaction] {MLP\\{[16]}};
  
  \path (prevaction) edge[flow] (embedaction);
  \node (concatembeds) [pnode, below=of embedaction] {Concat+FC\\ {[32]}\vphantom{[6]}};
  \path (embedaction) edge[flow] (concatembeds);

  \node (plus) [pnode, below=of preconv] {Add\\ {[64,64,32]}};
  \path (preconv) edge[flow] (plus);
  \path (concatembeds) edge[flow] (plus);

  \coordinate (condition_bottom) at ($ (plus.south)!0.5!(concatembeds.south) $);

  \mytrap{stridedconv}{downnode,below=of condition_bottom}{Conv 4x4\\{(stride=2)}\\{[$ \cdot $/2,$ \cdot $/2,32]}};
  \node (downsampling) [nonode,left=0.0cm of stridedconv] {$3\,\times$};

  \path (plus) edge[flow] (stridedconv);

  \node (resstack) [pnode, below= of stridedconv] {ResBlock 3x3\\{[8,8,32]}};
  \node (manyblocks) [nonode, left=0.0cm of resstack] {$8\,\times$};
  \path (stridedconv) edge[flow] (resstack);
  \node (prelstm) [pnode, below= of resstack] {Flatten+FC \\ {[256]}};
  \path (resstack) edge[flow] (prelstm);
  \node (lstm) [pnode, below= of prelstm] {LSTM\\ {[256]}};
  \path (prelstm) edge[flow] (lstm);
  \node (head) [pnode, below= of lstm] {Decoder\\\textnormal{(see \fig{autoregressive_head})}};

  \node (past) [nonode,left= of lstm] {Last state};
  \node (future) [nonode,right=of lstm] {Next state};
  \path (past) edge[flow] (lstm);
  \path (lstm) edge[flow] (future);
  \path (lstm) edge[flow] (head);

  \mytrap{output}{onode,below=of head,double copy shadow={shadow xshift=1ex,shadow yshift=0.6ex}}{$ a_{t + 1} $\\{[1]}};
  \path (head) edge[flow] (output);
\end{tikzpicture}
    \caption{The architecture of the \textbf{policy network} for a single step. \texttt{FC} refers to a fully-connected layer, \texttt{MLP} is a multilayer perceptron, \texttt{Conv} is a convolutional layer and \texttt{ResBlock} is a residual block. We give the dimensions of the output tensors in the square brackets. ReLU activations between the layers have been omitted for brevity.}
    \label{fig:policy}
\end{figure}

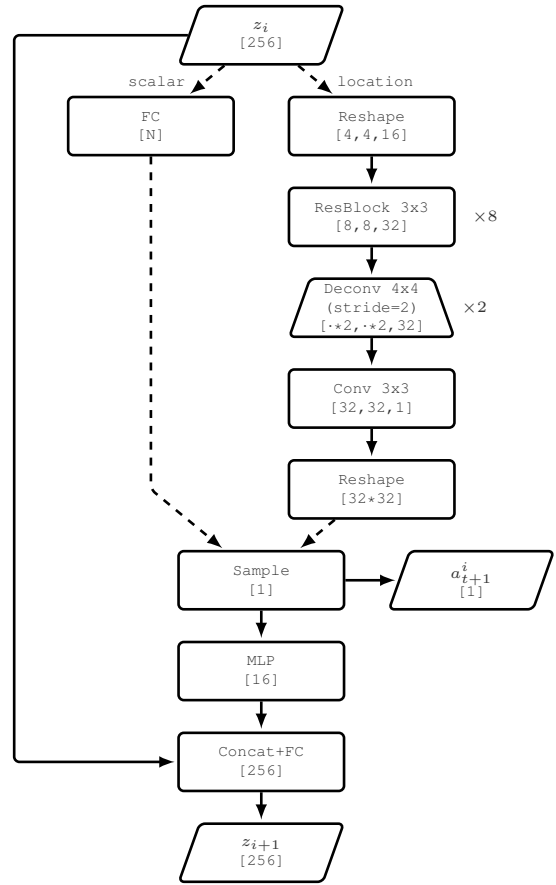
\begin{figure}
    \centering
    \begin{tikzpicture}[
  scale=0.3,
  black!100,
  text=black!80,
  node distance=0.4cm and 0.7cm,
  fnode/.style={
    align=center,
    rectangle,minimum height=22pt,minimum width=2.2cm,rounded corners=2pt,
    inner sep=3pt,
    line width=1pt,
    fill=pnodefill,draw=pnodedraw,
    font=\ttfamily\tiny},
  nonode/.style={
    align=center,
    font=\ttfamily\tiny},
  dnode/.style={
    fnode,trapezium,trapezium left angle=70, trapezium right angle=110,trapezium stretches},
  inode/.style={
    dnode,fill=inodefill,draw=inodedraw},
  downnode/.style={
    fnode,trapezium,trapezium left angle=110, trapezium right angle=110,trapezium stretches},
  upnode/.style={
    fnode,trapezium,trapezium left angle=70,trapezium right angle=70,trapezium stretches},
  pnode/.style={
    fnode,fill=pnodefill,draw=pnodedraw},
  pnode2/.style={
    fnode,fill=pnodefill!50!white,draw=pnodedraw!50!white},
  onode/.style={
    dnode,fill=onodefill,draw=onodedraw},
  smnode/.style={
    mnode,fill=mnodefill,draw=mnodedraw},
  flow/.style={
    -latex,shorten >=1pt,line width=1pt,line cap=rectangle,rounded corners=2pt,draw=pnodedraw,draw==\#1},
  flow2/.style={
    flow,draw=pnodedraw!50!white}]
  \newcommand{\mytrap}[3]{%
    \node (#1) [anchor=center,#2] {\phantom{$ C_t $}\\\phantom{[64,64,1]}}; \node[anchor=center,align=center,font=\tiny\ttfamily] at (#1.center) {#3};
  }

  \node (reshape1) [pnode] {Reshape\\ {[4,4,16]}};
  \node (resstack) [pnode, below= of reshape1] {ResBlock 3x3\\ {[8,8,32]}};
  \node (resblocktimes) [nonode,right=0.1cm of resstack] {$\times 8$};
  \path (reshape1) edge[flow] (resstack);
  \mytrap{deconv}{upnode, below= of resstack}{Deconv 4x4\\{(stride=2)}\\{[$ \cdot $*2,$ \cdot $*2,32]}};
  \node (deconvtimes) [nonode,right=0.1cm of deconv] {$\times 2$};
  \path (resstack) edge[flow] (deconv);
  \node (lastconv) [pnode, below= of deconv] {Conv 3x3\\ {[32,32,1]}};
  \path (deconv) edge[flow] (lastconv);
  \node (reshape2) [pnode, below= of lastconv] {Reshape\\ {[32*32]}};
  \path (lastconv) edge[flow] (reshape2);

  \node (fc) [pnode, left=of reshape1] {FC\\{[N]}};

  \coordinate (branches_top) at ($ (fc.north)!0.5!(reshape1.north) $);
  \mytrap{input}{dnode,above=of branches_top}{$ z_i $\\{[256]}};
  \path (input) edge[flow,dashed] node[nonode,right=4pt] {location} (reshape1);
  \path (input) edge[flow,dashed] node[nonode,left=4pt] {scalar} (fc);

  \coordinate (scalar_bottom) at ($ (reshape2.south) - (reshape1.south) + (fc.south) $);
  \coordinate (scalar_reshape_center) at ($ (reshape2.center) - (reshape1.south) + (fc.south) $);
  \coordinate (branches_bottom) at ($ (scalar_bottom)!0.5!(reshape2.south) $);

  \node (sample) [pnode,below=of branches_bottom] {Sample\\ {[1]}};
  \path (reshape2) edge[flow,dashed] (sample);
  \draw [flow,dashed] (fc) -- (scalar_reshape_center) -- (sample);
  \mytrap{action}{dnode,right=of sample}{$ a_{t + 1}^i $\\{[1]}};
  \path (sample) edge[flow] (action);
  \node (embed) [pnode,below=of sample] {MLP\\{[16]}};
  \path (sample) edge[flow] (embed);
  \node (final) [pnode,below=of embed] {Concat+FC\\{[256]}};
  \path (embed) edge[flow] (final);
  \mytrap{output}{onode,below=of final}{$ z_{i + 1} $\\{[256]}};
  \path (final) edge[flow] (output);

  \path[flow] (input.west) -| ($ (fc.west) + (fc.east) - (reshape1.west) $) |- (final.west);
\end{tikzpicture}
    \caption{The architecture of the \textbf{autoregressive decoder} for sampling an element $ a_{t + 1}^i $ of the action tuple. The initial hidden vector $ z_0 $ is provided by an upstream LSTM. Depending on the type of the subaction to be sampled, we use either the \texttt{scalar} or the \texttt{location} branch of the diagram.}
    \label{fig:autoregressive_head}
\end{figure}

The policy network (shown in \fig{policy}) takes the observation (\ie, the current state of the canvas $ C_t $) and conditions it on a tuple corresponding to the last performed action $ a_t $. The resulting features are then downsampled to a lower-dimensional spatial resolution by means of strided convolutions and passed through a stack of ResNet blocks \cite{He16} followed by a fully-connected layer. This yields an embedding which we feed into an LSTM \cite{Hochreiter97}. The LSTM produces a hidden vector $ z_0 $ serving as a seed for the action sampling procedure described below.

In order to obtain $ a_{t + 1} $, we employ an \emph{autoregressive decoder} depicted in \fig{autoregressive_head}. Each component $ a_{t + 1}^i $ is sampled from a categorical distribution whose parameters are computed as a function of $ z_i $. We use two kinds of functions depending on whether $ a_{t + 1}^i $ corresponds to a scalar (\eg, brush size) or to a spatial location (\eg, a control point of a Bézier curve). In the scalar case, $ z_i $ is transformed by a fully-connected layer, otherwise we process it using several ResNet blocks followed by a series of transpose convolutions and a final convolution. After $ a_{t + 1}^i $ is sampled, we obtain an updated hidden vector $ z_{i + 1} $ by embedding $ a_{t + 1}^i $ into a 16-dimensional code and combining it with $ z_i $. The procedure is repeated until the entire action tuple has been generated.

For the discriminator network, we use a conventional architecture similar to DCGAN \cite{Radford15}.

\section{Training Details}

Following standard practice in the GAN literature, we optimize the discriminator objective using Adam \cite{Kingma14} with a learning rate of $ 10^{-4} $ and $ \beta_1 $ set to $ 0.5 $. For generator training, we employ  population-based exploration of hyperparameters (PBT) \cite{Jaderberg17} to find values for the entropy loss coefficient and learning rate of the policy learner. A population contains 12 training instances with each instance running 64 CPU actor jobs and 2 GPU jobs (1 for the policy learner and 1 for the discriminator learner). We assume that discriminator scores are compatible across different instances and use them as a measure of fitness in the exploitation phase of PBT.

The batch size is set to 64 on both the policy learner and discriminator learner. The generated data is sampled uniformly from a replay buffer with a capacity of 20 batches. 

\end{document}